\documentclass[letterpaper, 10 pt, conference]{ieeeconf}  

\IEEEoverridecommandlockouts                              

\usepackage{graphics} 
\usepackage{amsmath} 
\usepackage{amssymb}  
\usepackage[noadjust]{cite}
\usepackage{xcolor}
\usepackage{multirow}
\usepackage{caption}
\usepackage{graphicx}
\usepackage[hidelinks]{hyperref}
\usepackage[all]{hypcap}
\usepackage{booktabs}

\colorlet{red}{black}
\colorlet{magenta}{black}

\title{\LARGE \bf
Distilling Privileged Control Barrier Functions into RGB-Only Safety Filters for Dynamic Visual Navigation
}

\author{Seungyeon Yoo$^{*}$, Gawon Lee$^{*}$, Seungwoo Jung$^{*}$, Inkyu Jang, and H. Jin Kim 
\thanks{$^{*}$Equal contribution.}
\thanks{Seungyeon Yoo, Gawon Lee, Seungwoo Jung, Inkyu Jang, and H. Jin Kim are with the Department of Aerospace Engineering, Seoul National University, Seoul, Republic of Korea
        {\tt\small syeon.yoo@snu.ac.kr, lgw1997@snu.ac.kr, tmddn833@snu.ac.kr, janginkyu@berkeley.edu, hjinkim@snu.ac.kr}}%
}

\begin{document}

\maketitle
\thispagestyle{empty}
\pagestyle{empty}

\begin{abstract}

RGB-only end-to-end visual navigation policies remain vulnerable to collisions in real-world dynamic environments, motivating a dedicated safety layer.
Existing visual Control Barrier Function (CBF) approaches seek to provide safety from RGB observations, but often rely on real-time rendering or explicit scene reconstruction and are primarily designed for static scenes, limiting their practicality for onboard deployment.
We propose a teacher-student visual distillation framework that transfers the safety behavior of a privileged CBF teacher to an RGB-only student filter for dynamic environments.
The student maps a short RGB history, robot velocity, and a nominal control action directly to a safe action, while the teacher uses ground-truth robot and obstacle states in a real-to-sim dynamic Gaussian Splatting environment.
To reduce the teacher-student information gap, the teacher constructs safety constraints only from obstacles observable within the student’s RGB history.
It also accounts for obstacle-velocity uncertainty to improve robustness to motion variations, while action augmentation exposes the student to diverse safe and unsafe nominal actions to better capture the safety boundary.
At deployment, the student requires only RGB observations and robot velocity, without explicit 3D reconstruction or online rendering. Experiments show that the proposed method outperforms visual CBF baselines and improves the safety of RGB-based navigation policies under dynamic obstacle motion.
Project page: \href{https://syeon-yoo.github.io/distill-cbf-site/}{\texttt{https://syeon-yoo.github.io/distill-cbf-site/}}.

\end{abstract}

\section{Introduction}

RGB-only end-to-end visual navigation policies which directly map camera observations to control actions offer several practical advantages for real-world deployment. By relying only on an RGB camera, they reduce hardware complexity, cost, and weight. They also remove the need for explicit mapping or reconstruction in the control loop, enabling low-latency visual feedback. Despite continued advances in end-to-end navigation \cite{8877728, readygo, shah2023vint}, however, such policies remain susceptible to collisions, particularly in dynamic environments. This lack of safety is a major obstacle to their deployment and motivates a dedicated safety layer.

Designing such a safety layer without sacrificing the simplicity of RGB-only navigation is challenging. Conventional safety mechanisms often rely on robot localization, obstacle-state estimation, depth sensing, or explicit 3D scene representations. Introducing these components at deployment would undermine the main benefit of an end-to-end RGB-only policy \cite{10802694}. The difficulty is further amplified in dynamic environments, where safe control requires reasoning not only about obstacle geometry but also about obstacle motion and its uncertainty, all from RGB observations within a limited field of view \cite{10172259, 10273593}. Thus, for RGB-only visual navigation, it is desirable for a safety filter to operate directly from visual observations in real time, without relying on explicit scene reconstruction and obstacle-state estimation at deployment.

Control Barrier Functions (CBFs) provide a systematic way to improve navigation safety by constraining the control input \cite{8796030}. Given a set of safe states, CBFs admit only control inputs that keep the system within the set, and a CBF-based safety filter minimally modifies a nominal action to satisfy this constraint. Recent visual CBF approaches \cite{nerf-cbf, safer-splat} extend this idea to image-based observations. However, they assume static scenes, and their safety evaluation requires an 3D scene representation at deployment, such as real-time depth rendering from a Neural Radiance Field (NeRF) \cite{mildenhall2020nerf} or a 3D Gaussian Splatting (GS) map \cite{kerbl3Dgaussians}. These limitations hinder their real-time onboard operations in dynamic environments.

\begin{figure}[t]
    \centering
    \includegraphics[width=0.96\linewidth]{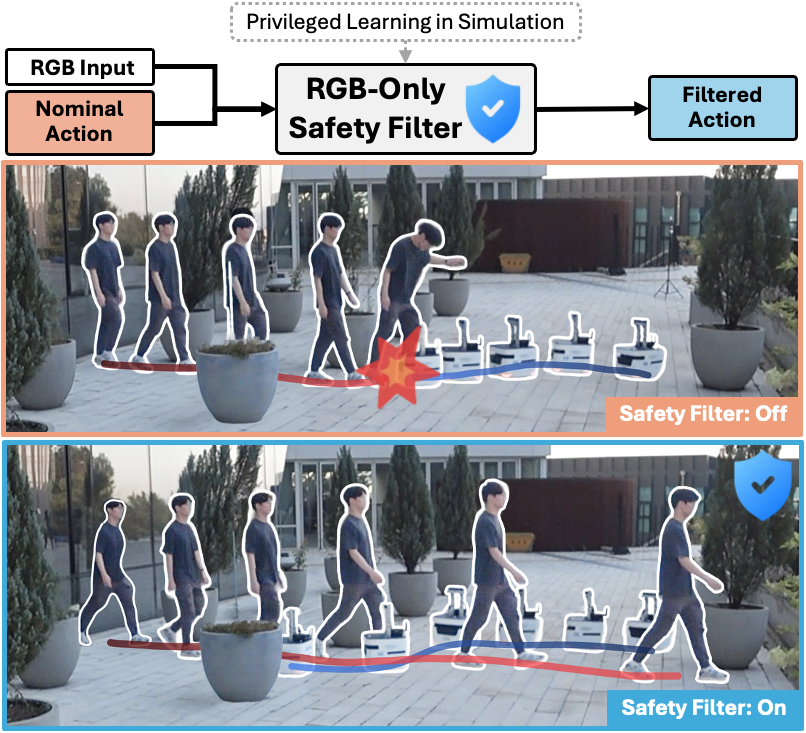}
    \caption{
    \textit{Real-world navigation example of the proposed RGB-only safety filter.} We distill a privileged state-based CBF teacher into an RGB-only student filter in a real-to-sim simulation with moving obstacles. In real-world dynamic environments, the proposed filter improves both collision avoidance and task success of nominal visual navigation policies, without requiring a 3D scene representation, robot localization, or obstacle state estimation at deployment.
    }
    \label{fig:thumbnail}
    \vspace{-0.5cm}
\end{figure}

This work proposes a different approach: rather than evaluating a CBF from a 3D scene representation at deployment, we use a privileged state-based CBF only during training and distill its safety behavior into an RGB-only student safety filter.
As illustrated in Fig.~\ref{fig:thumbnail}, \textcolor{magenta}{the student maps a nominal action directly to a safe action using RGB inputs}, while the privileged teacher uses ground-truth robot and obstacle states available in simulation.
To reduce the teacher-student information gap, the teacher considers only obstacles appearing in the short history of RGB inputs provided to the student.
It further accounts for uncertainty in obstacle velocities to improve robustness to changes in obstacle motion. Moreover, we augment the training data with multiple teacher-labeled candidate actions for each observation to encourage the student to learn the boundary between safe and unsafe actions.
Training data are generated in a real-to-sim dynamic GS simulation of each target environment \cite{readygo}, providing photorealistic observations and privileged states across diverse dynamic scenarios.
At deployment, the student filters nominal actions using only RGB observations and robot velocity, without requiring a 3D scene representation, robot localization, or obstacle state estimation.

Our contributions can be summarized as follows:
\begin{itemize}
    \item A privileged CBF-to-RGB distillation framework enabling RGB-only safety filtering in dynamic environments without a 3D scene representation, robot localization, or obstacle state estimation at deployment.

    \item A privileged CBF teacher design that reduces the teacher-student information gap by considering only obstacles appearing in the student's short RGB input sequence and accounts for uncertainty in obstacle velocities for robustness to unpredictable obstacle motion.

    \item Experiments with multiple nominal policies showing that our filter outperforms prior visual CBF methods in simulation and reduces collisions in real-world dynamic environments.
\end{itemize}

\section{Related Work}

\subsection{RGB-Only Visual Navigation}

\textcolor{magenta}{RGB-only end-to-end visual navigation policies directly map visual observations to actions and are commonly trained in simulation using imitation or reinforcement learning \cite{9359345, 9341049, 10542210, 8877728}.} Despite efforts to improve performance through domain randomization \cite{8877728} or cross-modal supervision \cite{9341049, 10542210}, their real-world applicability remains limited by the sim-to-real gap and the absence of safety considerations.

\textcolor{magenta}{One alternative direction is general navigation models (GNMs) \cite{shah2023vint, shah2022gnm, 10610665}, which leverage large real-world cross-embodiment datasets to improve zero-shot transfer across environments and robots. Another direction is GS-based real-to-sim pipelines \cite{readygo, 11020756, 10937041, 11095066}, which build a digital twin from a video capture and generate large-scale photorealistic navigation data, including for dynamic scenes \cite{readygo, 11095066}. These approaches improve the real-world applicability of RGB-only navigation policies by reducing the gap between training and deployment. However, safety around moving obstacles remains insufficiently addressed, motivating a dedicated safety layer for RGB-only visual navigation.}

\subsection{Visual CBF-Based Safety Filter}

When safety filters are deployed on real robots, the system state---obstacle shapes, poses, and velocities---is not available, and safety must be assessed from onboard perception alone.
\textcolor{magenta}{Prior methods recover this state at test time, either from a scene representation or from depth.}
NeRF-CBF~\cite{nerf-cbf} certifies each candidate action by rendering the predicted next observation with a NeRF~\cite{mildenhall2020nerf}, and SAFER-Splat~\cite{safer-splat} constructs its CBF from the Gaussian primitives of an online-built splat~\cite{kerbl3Dgaussians}.
Both assume a static world, since relaxing this would require remapping the scene faster than obstacles move.
On the depth side, V-CBF~\cite{vcbf} learns an image-space barrier from RGB-D input but assumes stationary obstacles and requires segmentation labels, while Depth-CBF~\cite{depthcbf} fits a local quadratic barrier to the point cloud without training but keeps no history and hence no notion of obstacle velocity.
Both presuppose a depth sensor.
Instead, we distill a privileged CBF-based filter into one that runs on RGB alone, requiring no depth sensor, scene model, rendering/mapping at inference while accounting for obstacle motion.

\textcolor{magenta}{
\subsection{Learning from Privileged Teachers}
Privileged learning and teacher-student approaches exploit information available only during training to improve policies that operate under partial observations at deployment~\cite{vapnik2009new,chen2020learning,kumar2021rma,monaci2022dipcan}.
Such approaches have been widely used for policy learning and perception, in which a privileged state-based teacher supervises a student acting on raw sensors~\cite{chen2020learning,kumar2021rma,monaci2022dipcan}, but rarely for distilling the corrective behavior of a model-based safety filter.
Our work specifically distills a privileged state-based CBF into an RGB-only safety filter, while explicitly addressing the teacher-student information gap induced by partial visual observations.
}

\section{Preliminaries: Control Barrier Functions}
\label{sec:prelim_cbf}

{\color{magenta}
Consider a control-affine system
\begin{equation}
    \dot{\mathbf{x}} = f(\mathbf{x}) + g(\mathbf{x})\,\mathbf{u},
    \qquad
    \mathbf{x}\in\mathcal{X},\;\;
    \mathbf{u}\in\mathcal{U},
    \label{eq:affine}
\end{equation}
where $\mathbf{x}$ and $\mathbf{u}$ denote the state and control input, respectively.
Safety is described by a continuously differentiable function $h:\mathcal{X}\rightarrow\mathbb{R}$, with the safe set $\mathcal{C}:=\{\mathbf{x}\mid h(\mathbf{x})\ge0\}$.
Thus, $h>0$ indicates a safe state, while $h=0$ defines the safety boundary.

A control barrier function (CBF) keeps the system inside $\mathcal{C}$ by constraining how quickly $h$ can decrease.
For an extended class-$\mathcal{K}$ function $\alpha(\cdot)$, a safe control input satisfies
\begin{equation}
    L_f h(\mathbf{x}) + L_g h(\mathbf{x})\,\mathbf{u}
    \ge -\alpha\!\left(h(\mathbf{x})\right).
    \label{eq:cbf_cond}
\end{equation}
Under standard regularity conditions, enforcing this inequality renders
$\mathcal{C}$ forward invariant \cite{8796030}.

Following \cite{ames2016control}, the CBF condition \eqref{eq:cbf_cond} is affine in the control input, it can be incorporated into a quadratic program to minimally modify a reference control input:
\begin{equation}
\begin{aligned}
    \mathbf{u}^{\star}
    &= \arg\min_{\mathbf{u}\in\mathcal{U}}
      \left\|\mathbf{u}-\mathbf{u}_{\mathrm{ref}}\right\|_{2}^{2} \\
    \mathrm{s.t.}\quad
    &L_f h + L_g h\,\mathbf{u}
      \ge -\alpha(h).
\end{aligned}
\label{eq:cbfqp}
\end{equation}
This is a small quadratic program (CBF-QP) that returns the reference input unchanged whenever it is already safe and otherwise makes the smallest correction that restores safety.
Multiple obstacles can be handled by stacking one CBF constraint per barrier $h_i$, provided that the resulting QP remains feasible.
The reference controller and the safety constraint are fully decoupled, which is what allows us to wrap an arbitrary visual navigation policy with a CBF-QP.
}

\section{Problem Formulation}
\label{sec:problem_formulation}
 
We consider a wheeled mobile robot equipped with a single forward-facing RGB camera, navigating a fixed target environment that contains static obstacles and a moving human.
The robot is modeled as a unicycle with state $\mathbf{x}^{\mathrm{r}} = (p_x,\, p_y,\, \psi,\, v)$, i.e., planar position, heading, and forward speed, driven by the control input $\mathbf{u} = (a,\, \omega)$ of forward acceleration and yaw rate:
\begin{equation}
    \dot{p}_x = v\cos\psi,\quad
    \dot{p}_y = v\sin\psi,\quad
    \dot{\psi} = \omega,\quad
    \dot{v} = a,
    \label{eq:unicycle}
\end{equation}
which is control-affine and hence of the form \eqref{eq:affine}.
The environment is described by a set of static obstacles $\mathcal{O}$ and the human state $\mathbf{x}^{\mathrm{h}} = (\mathbf{p}^{\mathrm{h}},\, \mathbf{v}^{\mathrm{h}})$.
We collect these into the \emph{privileged state} $\mathbf{x} = (\mathbf{x}^{\mathrm{r}},\, \mathbf{x}^{\mathrm{h}},\, \mathcal{O})$ and define the safe set $\mathcal{C} = \{\mathbf{x} \mid h(\mathbf{x}) \ge 0\}$, where $h$ is positive whenever the robot keeps a minimum clearance from both the static obstacles and the human.
At deployment the filter can rely only on onboard quantities: a short history of RGB frames $o_t = (I_{t-2\Delta},\, I_{t-\Delta},\, I_t)$ spaced $\Delta$ control steps apart, the nominal command $\mathbf{u}^{\mathrm{nom}}_t$, and the forward speed $v_t$ from the wheel encoder.
 
\textbf{Nominal policy.}
The robot is driven by an RGB-only navigation policy $\pi^{\mathrm{nom}}$ that, at each control step $t$, maps the current image and the robot speed to a reference command $\mathbf{u}^{\mathrm{nom}}_t = \pi^{\mathrm{nom}}(o_t, v_t)$.
We make no assumption on how $\pi^{\mathrm{nom}}$ was obtained and treat it as a black box that provides no safety guarantee of its own.

\textbf{Privileged safety filter.}
If the privileged state were available, the CBF-QP \eqref{eq:cbfqp} with $\mathbf{u}_{\mathrm{ref}} = \mathbf{u}^{\mathrm{nom}}_t$ would yield a minimally invasive safe command
\begin{equation}
    \mathbf{u}^{\mathrm{safe}}_t = \mathcal{F}\big(\mathbf{x}_t,\, \mathbf{u}^{\mathrm{nom}}_t\big),
    \label{eq:teacher_filter}
\end{equation}
which coincides with $\mathbf{u}^{\mathrm{nom}}_t$ away from $\partial\mathcal{C}$ and overrides it only when needed to keep the system in $\mathcal{C}$.
However, evaluating $\mathcal{F}$ requires the robot pose, the position and velocity of the human, and the scene geometry.

\textbf{Objective.}
Our goal is to learn an \emph{RGB-only safety filter} $\mathcal{F}_\theta : (o_t,\,v_t,\,\mathbf{u}^{\mathrm{nom}}_t) \mapsto \hat{\mathbf{u}}^{\mathrm{safe}}_t$ that reproduces the behavior of the privileged filter without access to $\mathbf{x}_t$:
\begin{equation}
    \min_\theta\;
    \mathbb{E}_{\mathcal{D}}
    \Big\lVert
        \mathcal{F}_\theta \big(o_t,\,v_t,\,\mathbf{u}^{\mathrm{nom}}_t\big)
        - \mathcal{F}\big(\mathbf{x}_t,\, \mathbf{u}^{\mathrm{nom}}_t\big)
    \Big\rVert_2^2,
    \label{eq:problem_objective}
\end{equation}
where $(o_t,\, v_t,\, \mathbf{x}_t,\, \mathbf{u}^{\mathrm{nom}}_t) \sim \mathcal{D}$ is the distribution of states and observations visited when the nominal policies operate in the target environment.
We use a photorealistic digital twin of the target environment can be built from a single camera capture, in which the privileged state is available and dynamic scenarios can be generated at scale.

Two properties make the problem non-trivial.
First, to preserve the low-latency benefit of the end-to-end nominal policy, $\mathcal{F}_\theta$ must filter actions directly from $o_t$ and $v_t$, without estimating obstacle geometry at deployment.
Second, \eqref{eq:problem_objective} is only well-posed if the teacher's decisions are inferable from $o_t$.
In other words, $\mathcal{F}$ must depend on the privileged state only through quantities that are observable from a short RGB history, such as the relative position and motion of the human when it is in view.
These two properties drive the distillation framework and the teacher design in Section~\ref{sec:method}.

\begin{figure}[t]
  \centering
  \includegraphics[width=1.0\linewidth]{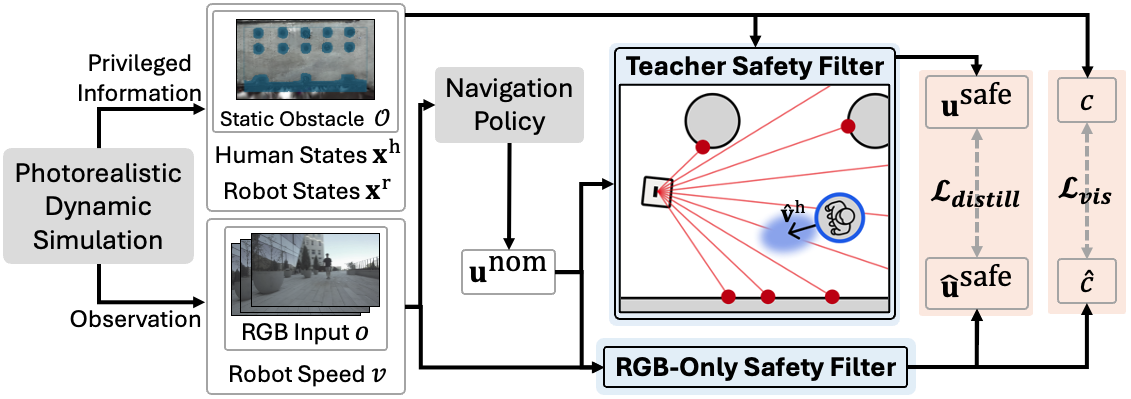}
    \caption{
    \textit{Training pipeline.}
    We train the RGB-only student safety filter by distilling a privileged teacher safety filter.
    The teacher draws its privileged state from the simulation, while the student learns to reproduce the teacher's safe action from the observations available at deployment alone.
    }
  \label{fig:training-pipeline}
  \vspace{-0.6cm}
\end{figure}

\section{RGB-Only Safety Filter Distillation} \label{sec:method}

To distill a CBF-based safety filter which relies on ground-truth states into a RGB-only safety filter, we design two main components, as shown in Fig.~\ref{fig:training-pipeline}.
The first component is a teacher CBF-based safety filter which maps nominal actions produced by RGB-only navigation policies to safe actions, using privileged information provided from a simulator.
The second component is the student RGB-only safety filter.
The student maps a stack of RGB observations and a nominal action to a safe action by distilling the teacher inside the photorealistic simulator~\cite{readygo}, which allows us to deploy the student in the real environments without privileged information and with multiple nominal policies.

\subsection{Privileged State-Based CBF Teacher}
\label{sec:method_teacher}

The teacher could read the exact simulator state, but we restrict it to what the student can in principle infer from $o_t$, so that the distillation objective~\eqref{eq:problem_objective} is well posed.
Static geometry enters as the nearest raycast hit per bearing across the robot's field of view, giving
$N$ boundary points treated as discs. 
The human enters only if it was inside the field of view at one of the three steps $\{t,\,t-\Delta,\,t-2\Delta\}$ of $o_t$.
With $t - \tau$ ($\tau \in \{ 0, \Delta, 2\Delta \}$) the most recent such step, the teacher uses the finite-difference velocity $\hat{\mathbf{v}}^{\mathrm{h}}$ at that step and the extrapolated position $\hat{\mathbf{p}}^{\mathrm{h}} = \mathbf{p}^{\mathrm{h}}(t-\tau) + \tau \hat{\mathbf{v}}^{\mathrm{h}}$. 
The teacher never sees the human's current velocity, only a stale estimate of it, and this is the uncertainty the barrier must absorb.

\textbf{\textcolor{magenta}{Uncertainty-aware DP-CBF.}}
We build on the dynamic parabolic CBF (DP-CBF) \cite{park2026dpcbf}, which incorporates the relative velocity of a dynamic obstacle but assumes that velocity is known exactly and constant.
In our setting, the human's velocity is inferred from a short observation history and becomes stale when the human accelerates, turns, or leaves the field of view, so treating the estimate as exact can make the barrier overconfident.
Feeding the teacher the exact privileged velocity would avoid this but widen the information gap to the RGB-only student.
\textcolor{magenta}{We therefore augment the DP-CBF with an uncertainty-dependent tightening margin, while keeping the teacher consistent with what the student can infer from its observations.}

With $(\tilde v_{\text{rel},x},\tilde v_{\text{rel},y})$ the relative velocity in an obstacle's frame where the x-axis is aligned with the line-of-sight,
\begin{equation}
  h= \tilde{v}_{\text{rel},x} + \lambda(\mathbf{x})\,\tilde{v}_{\text{rel},y}^{2}+\mu(\mathbf{x}),
  \label{eq:dpcbf}
\end{equation}
where $\lambda,\mu > 0$ are clearance-dependent gains defined in \cite{park2026dpcbf}.
For the human, $h$ is evaluated at $\hat{\mathbf{v}}^{\mathrm{h}}$, which is in error whenever the human changes speed or heading within $\tau$.
We model the error as
$\boldsymbol{\varepsilon}\sim\mathcal N(\mathbf{0},\Sigma)$,
with an anisotropic covariance in the human's motion frame,
$\Sigma=\sigma_\parallel^{2} \hat{\mathbf{u}}\hat{\mathbf{u}}^{\!\top}+\sigma_\perp^{2}(I-\hat{\mathbf{u}}\hat{\mathbf{u}}^{\!\top}),
\hat{\mathbf{u}}={\hat{\mathbf{v}}^{\mathrm{h}}} / {\lVert\hat{\mathbf{v}}^{\mathrm{h}}\rVert}.$
\textcolor{magenta}{
We use an anisotropic covariance with $\sigma_\parallel>\sigma_\perp$, assuming that pedestrians are more likely to vary their speed than abruptly change heading. This emphasizes uncertainty along the current motion direction, encouraging the filter to yield to crossing pedestrians without uniformly enlarging the safety margin.}

\textcolor{magenta}{We propagate this velocity uncertainty to the barrier using a first-order approximation,}
$
\textcolor{magenta}{
\sigma_h^{2}
=
\nabla_{\hat{\mathbf{v}}}h\,
\Sigma\,
\nabla_{\hat{\mathbf{v}}}h^{\!\top},\;
\tilde h=h-\beta\,\sigma_h,
}$
\textcolor{magenta}{
where $\beta>0$ controls conservatism and $\beta\sigma_h$ provides a tightening margin proportional to the propagated barrier uncertainty.
Inspired by Gaussian chance-constraint tightening \cite{li2026ccmpccbf}, we use this as an uncertainty-aware margin rather than a probabilistic guarantee, with fixed $\Sigma$ to avoid excessive conservatism under prolonged occlusions.}

\textbf{Teacher safety filter.}
Over the \(N\) static barriers \eqref{eq:dpcbf} and, when present, the uncertainty-aware human barrier, the teacher solves at every step
\begin{equation}
  \begin{aligned}
    \mathbf{u}^{\mathrm{safe}}_t=\arg\min_{\mathbf{u} \in \mathcal{U}}\;&\lVert \mathbf{u}- \mathbf{u}^{\mathrm{nom}}_t\rVert_2^{2}\\
    \text{s.t.}\;&L_f h_i+L_g h_i\,\mathbf{u}\ge-\alpha h_i\quad\\
    \;&a_{\min}\le a \le a_{\max}, \;\lvert\omega\rvert\le\omega_{\max}
  \end{aligned}
  \label{eq:teacherqp}
\end{equation}
This QP is strictly convex, so $\mathbf{u}^{\mathrm{safe}}_t$ is unique and varies continuously with $(\mathbf{x}_t,\mathbf{u}^{\mathrm{nom}}_t)$, a well-posed regression target for~\eqref{eq:distill_loss}, and it involves no quantity the student cannot recover from three frames.

\subsection{RGB-Only Student Safety Filter}

\textbf{Datasets.}
To train the student RGB-only safety filter, we collect datasets using the teacher in the dynamic GS simulation~\cite{readygo}.
At every control step $t$ we log the camera frames $o_t$, the robot speed $v_t$, the nominal policy's action $\mathbf{u}_t^{\mathrm{nom}}$, and the filtered action $\mathbf{u}_t^{\mathrm{safe}}$.
To cover both the states the unfiltered policy visits and the states the filtered system actually occupies, we collect two sets of rollouts.
One set is nominal rollouts, where the robot executes $\mathbf{u}_t^{\mathrm{nom}}$, and the other is filtered rollouts, where $\mathbf{u}_t^{\mathrm{safe}}$ is executed.
We discard samples whose QP did not converge to an optimal solution.
As nominal policies, we use imitation learning~\cite{readygo}, reinforcement learning~\cite{sac}, and general navigation~\cite{shah2023vint} policies.

Next, we augment the collected samples by perturbing the executed action and relabeling the sample with the teacher safety filter output.
The teacher relabels the samples offline at each logged state with perturbed candidate actions, producing additional $(o_t, v_t, \mathbf{u}_t^{\mathrm{nom}}, \mathbf{u}_t^{\mathrm{safe}})$ pairs without any further simulation.
We draw perturbed action candidates from a mixture of a Gaussian centered around the logged action and uniform distribution over the action space.
In addition to these candidates, when any CBF constraint intersects the action space, we draw more perturbed actions from that intersection.
This augmentation broadens the candidate action space coverage at each visited state, thereby making student truly approximate the teacher.

\textcolor{magenta}{
\textbf{Architecture and Loss Function.}
The student maps the observations available onboard to the teacher's filtered action.
Each frame is encoded independently by a ten-layer residual CNN into a 20-dimensional feature vector, and a three-layer MLP maps the concatenated frame features and scalar inputs to the predicted safe action $\hat{\mathbf{u}}_t^{\text{safe}} = \text{MLP}(\text{CNN}(o_t), v_t, \mathbf{u}^{\mathrm{nom}}_t)$.
Because interventions are rare, an action loss alone admits a shortcut that passes the nominal action through without grounding the decision to intervene in perception.
We therefore add an auxiliary head $\hat{c}_t = \text{MLP}_{\text{vis}}(\text{CNN}(o_t))$ that predicts from the image features alone whether the human is inside the camera's field of view, with the label $c_t \in \{0,1\}$ computed in the simulator from the ground-truth human position and robot pose.
Since the head never sees the scalar inputs, its gradients encourage the CNN to represent human presence, and it is discarded at deployment.
With $\mathbf{u}^{\mathrm{safe}}_t$ the teacher's QP solution, the training objective is
\begin{gather} \label{eq:distill_loss}
    \mathcal{L}(\theta) = \frac{1}{|\mathcal{D}|} \sum_{\mathcal{D}} \left[ \mathcal{L}_{\mathrm{distill}} + \lambda_{\mathrm{vis}}\,\mathcal{L}_{\mathrm{vis}} \right], \\
    \mathcal{L}_{\mathrm{distill}} = \big\lVert \hat{\mathbf{u}}_t^{\text{safe}} - \mathbf{u}^{\mathrm{safe}}_t \big\rVert_2^2, \\
    \mathcal{L}_{\mathrm{vis}} = \mathrm{BCE}(\hat{c}_t, c_t),
\end{gather}
where $\lambda_{\mathrm{vis}} = 0.05$ and $\mathrm{BCE}$ is the binary cross-entropy on the head's logit.
}

\section{Simulation Experiments}

We first evaluate the distilled visual safety filter in the ReaDy-Go simulator \cite{readygo}, which renders the reconstructed scenes and dynamic human avatars with Gaussian splatting.
We test whether the filter, operating only on onboard observations, improves both the safety and the task performance of the nominal policies.

\subsection{Experimental Setup}

\begin{table*}[t]
    \centering
    \caption{
        \textit{Simulation results.}
        Each nominal--filter--scene combination is evaluated over 100 trials.
        Best values are in bold.
    }
    \vspace{-0.1cm}
    \label{tab:sim_results}
    \footnotesize
    \setlength{\tabcolsep}{3pt}
    \begin{tabular}{llccccccccccccccc}
    \toprule
    & & \multicolumn{5}{c}{\textit{Outside}} & \multicolumn{5}{c}{\textit{Lobby}} & \multicolumn{5}{c}{\textit{Library}} \\
    \cmidrule(lr){3-7} \cmidrule(lr){8-12} \cmidrule(lr){13-17}
    Nominal & Safety filter & SR$\uparrow$ & CFR$\uparrow$ & $\bar m\uparrow$ & $T$ & Failure cases$\downarrow$ & SR$\uparrow$ & CFR$\uparrow$ & $\bar m\uparrow$ & $T$ & Failure cases$\downarrow$ & SR$\uparrow$ & CFR$\uparrow$ & $\bar m\uparrow$ & $T$ & Failure cases$\downarrow$ \\
    & & (\%) & (\%) & (m) & (s) & S/D/TO (\%) & (\%) & (\%) & (m) & (s) & S/D/TO (\%) & (\%) & (\%) & (m) & (s) & S/D/TO (\%) \\
    \midrule
    \multirow{5}{*}{IL} & None & 63 & 64 & 0.26 & 10.2 & 17/19/1 & 65 & 65 & 0.30 & 8.4 & 4/31/0 & 56 & 56 & 0.19 & 9.6 & 25/19/0 \\
     & Teacher & 88 & 98 & 0.37 & 12.6 & 0/2/10 & 85 & 87 & 0.41 & 10.4 & 0/13/2 & 89 & 93 & 0.34 & 12.2 & 0/7/4 \\
    \cmidrule(lr){2-17}
     & SAFER-Splat & 72 & 81 & 0.30 & 11.0 & 6/13/9 & 71 & 72 & 0.33 & 8.8 & \textbf{0}/28/\textbf{1} & 67 & 72 & 0.25 & 10.4 & 14/14/5 \\
     & NeRF-CBF & 47 & 74 & 0.19 & 16.3 & 16/10/27 & 22 & 76 & 0.11 & 16.8 & 1/23/54 & 65 & 68 & 0.23 & 12.7 & 20/12/\textbf{3} \\
     & Ours & \textbf{87} & \textbf{93} & \textbf{0.35} & 12.2 & \textbf{2}/\textbf{5}/\textbf{6} & \textbf{78} & \textbf{82} & \textbf{0.37} & 10.4 & \textbf{0}/\textbf{18}/4 & \textbf{88} & \textbf{91} & \textbf{0.32} & 12.2 & \textbf{1}/\textbf{8}/\textbf{3} \\
    \midrule
    \multirow{5}{*}{RL} & None & 56 & 56 & 0.21 & 9.6 & 26/18/0 & 56 & 56 & 0.23 & 7.6 & 15/29/0 & 37 & 37 & 0.13 & 6.5 & 40/23/0 \\
     & Teacher & 93 & 95 & 0.43 & 12.1 & 1/4/2 & 85 & 88 & 0.48 & 10.2 & 1/11/3 & 72 & 90 & 0.23 & 10.1 & 0/10/18 \\
    \cmidrule(lr){2-17}
     & SAFER-Splat & 74 & 80 & 0.30 & 11.7 & \textbf{2}/18/6 & 61 & 67 & 0.27 & 7.6 & 4/29/6 & 53 & 55 & 0.19 & 7.5 & 21/24/2 \\
     & NeRF-CBF & 65 & 80 & 0.24 & 16.1 & 4/16/15 & 26 & 76 & 0.11 & 12.6 & 2/22/50 & 37 & 38 & 0.12 & 9.4 & 34/28/\textbf{1} \\
     & Ours & \textbf{87} & \textbf{90} & \textbf{0.38} & 11.5 & \textbf{2}/\textbf{8}/\textbf{3} & \textbf{82} & \textbf{84} & \textbf{0.46} & 10.6 & \textbf{0}/\textbf{16}/\textbf{2} & \textbf{65} & \textbf{82} & \textbf{0.21} & 10.1 & \textbf{7}/\textbf{11}/17 \\
    \midrule
    \multirow{5}{*}{ViNT} & None & 38 & 42 & 0.15 & 8.6 & 43/15/4 & 38 & 70 & 0.22 & 8.1 & 10/20/32 & 31 & 31 & 0.11 & 6.0 & 49/20/0 \\
     & Teacher & 70 & 95 & 0.29 & 10.3 & 2/3/25 & 49 & 96 & 0.31 & 10.9 & 0/4/47 & 60 & 81 & 0.21 & 10.5 & 7/12/21 \\
    \cmidrule(lr){2-17}
     & SAFER-Splat & 67 & 82 & 0.27 & 10.2 & 5/13/\textbf{15} & 44 & 85 & 0.25 & 8.9 & \textbf{0}/15/\textbf{41} & 58 & 60 & 0.20 & 7.8 & 16/24/\textbf{2} \\
     & NeRF-CBF & 25 & 60 & 0.12 & 14.1 & 27/13/35 & 14 & 82 & 0.09 & 15.4 & \textbf{0}/18/68 & 32 & 44 & 0.12 & 11.5 & 30/26/12 \\
     & Ours & \textbf{72} & \textbf{90} & \textbf{0.28} & 10.8 & \textbf{4}/\textbf{6}/18 & \textbf{47} & \textbf{89} & \textbf{0.28} & 13.1 & 3/\textbf{8}/42 & \textbf{60} & \textbf{74} & \textbf{0.21} & 9.9 & \textbf{9}/\textbf{17}/14 \\
    \bottomrule
    \end{tabular}
    \vspace{-0.5cm}
\end{table*}

\subsubsection{Task description}

The task is point-goal navigation, where the robot must travel from a start pose to a goal position while avoiding static obstacles and a moving human.
For each scene we sample random start-goal pairs in the free space.
A trial succeeds if the robot reaches the goal within the time limit of 50 seconds, and fails on a human collision, a static collision, or a timeout.
The trials, including the start-goal pairs and the human trajectory, are identical across all nominal policies and safety filters.

\subsubsection{Evaluation metrics}

We report the success rate (SR), the collision-free rate (CFR), the mean safety margin ($\bar m$), the average reaching time ($T$), and a breakdown of failure cases.
SR is the fraction of trials in which the robot reaches within 1\,m of the goal without a collision and within the time budget.
\textcolor{red}{CFR is the fraction of trials without any collision, regardless of whether the goal is reached.}
$\bar m$ measures minimum surface-to-surface distance per trial between the robot and any obstacles, averaged over all trials with every failed trial counted as zero clearance.
$T$ is the mean time from start to goal over all trials, with failed trials assigned the maximum time limit.
\textcolor{red}{Failure cases are reported as the percentage of trials ending in a static collision (S), a collision with the human (D), or a timeout (TO).}

\subsubsection{Nominal visual navigation policies}
\begin{itemize}
    \item \textbf{IL}:
    We train an imitation learning policy following the ReaDy-Go pipeline~\cite{readygo}.
    We roll out an expert planner in the simulator and train the policy with behavior cloning.

    \item \textbf{RL}:
    We train a per-scene policy with Soft Actor-Critic (SAC,~\cite{sac, stable-baselines3}).
    The policy runs at 5\,Hz, holding each command for four steps of the 20\,Hz physics.
    The reward combines a sparse terminal success reward, a collision penalty, a small per-step time penalty, and potential-based goal-progression reward.

    \item \textbf{ViNT~\cite{shah2023vint}}:
    ViNT is a GNM trained on large-scale, cross-embodiment data.
    We use the public checkpoint without fine-tuning.
    Since ViNT is image-goal conditioned while our scenarios specify goal positions, we capture a goal image from the scene at the goal position, oriented along the start-to-goal direction.
    Among the five waypoint outputs, we map the third to $(v, w)$, then convert the velocity $v$ to an acceleration reference $a$.
\end{itemize}

\subsubsection{Visual safety filter baselines}
\begin{itemize}
    \item \textbf{SAFER-Splat~\cite{safer-splat}}:
    We adapt SAFER-Splat, which operates directly on Gaussian primitives, and provide it with the same 3D Gaussians in the simulated environment.
    At every step, the Gaussians of the human are additionally inserted into its constraint set, giving it ground-truth knowledge of the human.
    The original formulation assumes double-integrator dynamics, and we rewrite its constraints in terms of the unicycle's acceleration and angular-velocity commands.

    \item \textbf{NeRF-CBF~\cite{nerf-cbf}}:
    We train NICE-SLAM~\cite{nice-slam} per scene on RGB-D data rendered from the simulator by densely sweeping the navigable region at the robot's camera height.
    Random perturbations of the nominal command are scored by predicting the depth observation at the resulting next state.
    Because the moving human is absent from the static map, its depth is rendered from the ground-truth pose and composited into the barrier, so the baseline receives the same human information as the other methods.
\end{itemize}
Both baselines use the same nominal policies, scenarios, dynamics, and success and collision criteria as ours, and unlike our filter they observe the human at its ground-truth pose at test time, as Gaussians for SAFER-Splat and as rendered depth for NeRF-CBF.

\subsection{Implementation Details}
\begin{figure}[t]
    \centering
    \includegraphics[width=1.0\linewidth]{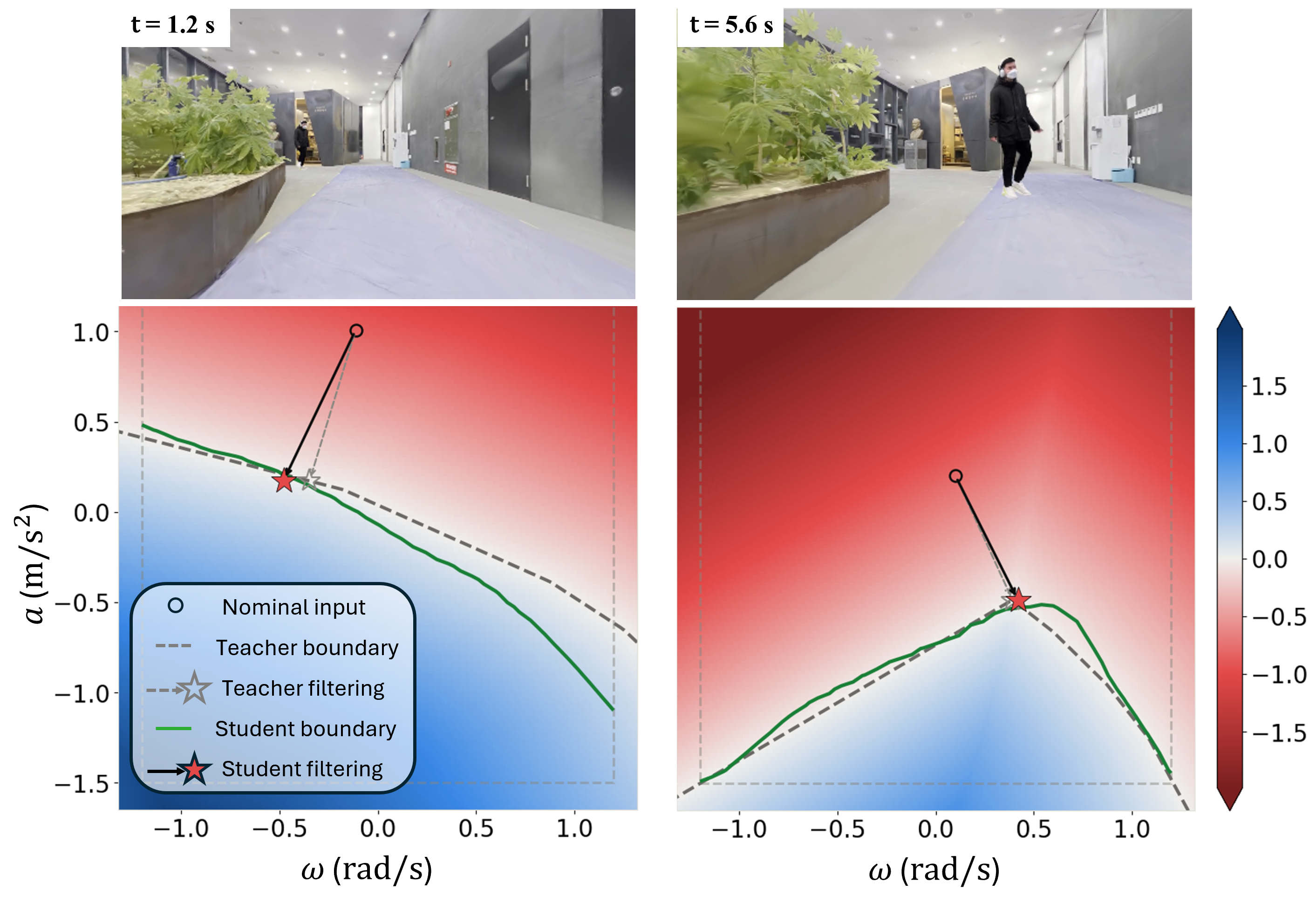}
    \vspace{-0.6cm}
    \caption{
    \textit{Reproducing the teacher's action-space filtering boundary.}
    \textcolor{red}{
    The top row shows the robot's observations in simulation, and the bottom row shows the corresponding action-space boundaries. 
    The color indicates the teacher CBF constraint margin, with the dashed gray curve denoting its zero-level boundary.
    The green curve denotes the student's intervention boundary.
    The results show that the RGB-only student's boundary closely matches the teacher's boundary.
    }
    }
    \label{fig:boundary_approx}
    \vspace{-0.6cm}
\end{figure}

We collect data in three scenes: \textit{Outside} (Fig.~\ref{fig:thumbnail}), \textit{Lobby} (Fig.~\ref{fig:boundary_approx}), and \textit{Library} (Fig.~\ref{fig:joystick_real_exp}(a)).
For each scene and nominal policy, we roll out 800 episodes with the unfiltered nominal policy and record the teacher's safe action as a counterfactual label at every transition where the teacher QP is solved.
To also cover the states the filtered system visits, we roll out another 800 episodes with the teacher filter.
The student for a scene is trained on the union of the six datasets (three nominal policies $\times$ two rollout types), which corresponds to the \emph{Mixed} setting in Table~\ref{tab:ablation}.
Each scene yields roughly 1.5M transitions, which grow to 12M after action augmentation.
Each student is trained on a single A6000 GPU.

\subsection{Navigation Results}

Table~\ref{tab:sim_results} shows that across all three scenes and nominal policies, the distilled filter raises the success \textcolor{red}{and collision-free rate,} and enlarges the safety margin over the unfiltered nominal policy.
Although the student never observes the privileged state used by the CBF-QP teacher, it closely tracks the teacher's success \textcolor{red}{and collision-free} rate while acting only on the RGB image stream, the nominal command, and the robot speed.
It also outperforms SAFER-Splat and NeRF-CBF in nearly every cell.
Both baselines observe the human at its ground-truth pose, but their barriers are built for static scenes and have no notion of its velocity.
They also depend on a 3D scene representation at test time and are therefore exposed to reconstruction artifacts.
NeRF-CBF additionally relies on a sampled search over perturbed actions and stops the robot when no candidate passes, which can forfeit progress near a moving human.
The distilled filter is supervised by a CBF-QP whose solution is unique and continuous, and at deployment it sees only RGB images, so it is free of these dependencies.
It also removes the need to build and maintain a 3D scene representation at deployment, since it is amortized into training.

\subsection{Filtering Boundary Reproduction by the Student}
\textcolor{red}{
We further examine whether the student reproduces the teacher's filtering boundary in the action space, as shown in Fig.~\ref{fig:boundary_approx}. The teacher boundary is given by the zero level set of the active CBF constraint margin, $\min_i \left(L_f h_i+L_g h_i\,\mathbf{u}+\alpha h_i\right)=0,$ which separates actions that satisfy the teacher's CBF constraints from those requiring intervention.
We define student's intervention boundary from the magnitude of its action correction. An action is considered unmodified when $\left\|\mathbf{u}_{\mathrm{safe}}-\mathbf{u}_{\mathrm{nom}}\right\|_2 < \epsilon,$ with $\epsilon=0.05$, and the corresponding $\epsilon$-level set defines the student intervention boundary.
}
\textcolor{red}{
Fig.~\ref{fig:boundary_approx} compares these boundaries at two time steps from the same simulated scenario. 
Despite having no access to the privileged state or explicit CBF constraints, the learned student boundary closely follows the teacher's constraint boundary as the scene changes. 
This indicates that the student distills not only individual safe-action labels, but also the local intervention structure induced by the privileged CBF teacher.
Specifically, in the left snapshot, the nearby obstacle shapes the boundary to reject a range of right-turning actions. 
In the right snapshot, the obstacle and pedestrian jointly constrain both steering directions, leaving only decelerating actions within the admissible region.
}

\subsection{Ablation Study}

\subsubsection{Effect of the \textcolor{magenta}{uncertainty-aware} margin}
To analyze how accounting for uncertainty in the human's velocity affects the teacher and the distilled student, we ablate the uncertainty-aware margin from the teacher.
The ablated teacher evaluates the barrier at the estimated human velocity alone, without \textcolor{magenta}{the uncertainty-aware margin}, and we distill a student from its labels.
Table~\ref{tab:human_uncertainty_agg} reports results averaged over all nominal policies and scenes.

The margin mainly changes where failures occur.
Removing it raises human collisions from 7.3\% to 12.1\% for the teacher and from 10.8\% to 15.3\% for the student, while static collisions and timeouts drop slightly.
Without the margin, the barrier trusts a stale velocity estimate, so a human who changes pace or heading after leaving the field of view can close the gap before the filter reacts.
The student reproduces this behavior without ever observing the human's velocity or its covariance, which indicates that the \textcolor{magenta}{uncertainty-aware} tightening margin is recoverable from the RGB history and is worth its modest cost in conservatism.

\begin{table}[t]
\centering
\caption{
\textit{Effect of the \textcolor{magenta}{uncertainty-aware} margin.} (w/o margin): the teacher without the \textcolor{magenta}{uncertainty-aware} margin on the human's velocity, and our filter distilled from its labels.
}
\vspace{-0.1cm}
\label{tab:human_uncertainty_agg}
\footnotesize
\setlength{\tabcolsep}{6pt}
\begin{tabular}{lccc}
\toprule
Safety filter & SR$\uparrow$ & CFR$\uparrow$ & Failure cases$\downarrow$ \\
 & (\%) & (\%) & S/D/TO (\%) \\
\midrule
None & 48.9 & 53.0 & 25.4/21.6/4.1 \\
\midrule
Teacher & \textbf{76.8} & \textbf{91.5} & 1.2/\textbf{7.3}/14.7 \\
Teacher (w/o \textcolor{red}{margin}) & 74.6 & 87.4 & \textbf{0.6}/12.1/\textbf{12.8} \\
\midrule
Ours & \textbf{74.0} & \textbf{86.1} & 3.1/\textbf{10.8}/12.1 \\
Ours (w/o \textcolor{red}{margin}) & 73.6 & 82.8 & \textbf{1.9}/15.3/\textbf{9.2} \\
\bottomrule
\end{tabular}
\vspace{-0.1cm}
\end{table}

\begin{table}[t]
\centering
\caption{
\textit{Effect of dataset aggregation and augmentation.}
\emph{Mixed}: The student is trained on the union of the datasets collected under all three nominal policies rather than its own nominal's.
\emph{Aug.}: Action augmentation of the CBF-QP labels.
}
\vspace{-0.1cm}
\label{tab:ablation}
\footnotesize\setlength{\tabcolsep}{5pt}
\begin{tabular}{ccccc}
\toprule
Mixed & Aug. & SR$\uparrow$ & CFR$\uparrow$ & Failure cases$\downarrow$ \\
 & & (\%) & (\%) & S/D/TO (\%) \\
\midrule
\multicolumn{2}{c}{Teacher} & 76.8 & 91.5 & 1.2/7.3/14.7 \\
\midrule
-- & -- & 72.2 & 83.6 & 4.7/11.7/\textbf{11.4} \\
-- & \checkmark & 69.7 & 81.6 & 4.7/13.8/11.9 \\
\checkmark & -- & 70.7 & 83.9 & 4.4/11.7/13.2 \\
\checkmark & \checkmark & \textbf{74.0} & \textbf{86.1} & \textbf{3.1}/\textbf{10.8}/12.1 \\
\bottomrule
\end{tabular}
\vspace{-0.5cm}
\end{table}

\begin{table*}[t]
    \centering
    \caption{
        \textit{Real-world navigation results.}
        Each nominal--filter--scene combination is evaluated over 10 trials.
    }
    \vspace{-0.2cm}
    \label{tab:real_results}
    \footnotesize
    \setlength{\tabcolsep}{3pt}
    \begin{tabular}{llccccccccccccccc}
    \toprule
    & & \multicolumn{5}{c}{\textit{Outside}} & \multicolumn{5}{c}{\textit{Lobby}} & \multicolumn{5}{c}{\textit{Library}} \\
    \cmidrule(lr){3-7} \cmidrule(lr){8-12} \cmidrule(lr){13-17}
    Nominal & Safety filter & SR$\uparrow$ & CFR$\uparrow$ & $\bar m\uparrow$ & $T$ & Failure cases$\downarrow$ & SR$\uparrow$ & CFR$\uparrow$ & $\bar m\uparrow$ & $T$ & Failure cases$\downarrow$ & SR$\uparrow$ & CFR$\uparrow$ & $\bar m\uparrow$ & $T$ & Failure cases$\downarrow$ \\
    & & (\%) & (\%) & (m) & (s) & S/D/TO (\%) & (\%) & (\%) & (m) & (s) & S/D/TO (\%) & (\%) & (\%) & (m) & (s) & S/D/TO (\%) \\
    \midrule
    \multirow{2}{*}{IL} & None & 30 & 40 & 0.13 & 41.1 & 50/10/10 & 70 & 70 & 0.32 & 26.0 & 0/30/0 & 50 & 50 & 0.20 & 33.4 & 0/50/0 \\
     & Ours & 50 & 80 & 0.07 & 36.8 & 20/0/30 & 90 & 90 & 0.41 & 28.1 & 0/10/0 & 80 & 80 & 0.34 & 29.0 & 0/20/0 \\
    \midrule
    \multirow{2}{*}{RL} & None & 50 & 50 & 0.18 & 34.9 & 20/30/0 & 20 & 20 & 0.05 & 43.3 & 50/30/0 & 50 & 50 & 0.13 & 32.8 & 10/40/0 \\
     & Ours & 80 & 80 & 0.30 & 22.8 & 10/10/0 & 60 & 80 & 0.24 & 33.9 & 0/20/20 & 70 & 70 & 0.33 & 30.0 & 0/30/0 \\
    \midrule
    \multirow{2}{*}{ViNT} & None & 40 & 40 & 0.16 & 35.3 & 20/40/0 & 20 & 30 & 0.10 & 42.1 & 50/20/10 & 40 & 40 & 0.23 & 33.3 & 20/40/0 \\
     & Ours & 70 & 80 & 0.27 & 23.4 & 20/0/10 & 50 & 70 & 0.31 & 32.3 & 20/10/20 & 70 & 80 & 0.34 & 25.6 & 20/0/10 \\
    \bottomrule
    \end{tabular}
    \vspace{-0.7cm}
\end{table*}

\begin{figure*}[t]
  \vspace{.4cm}

  \centering
  \includegraphics[width=0.89\linewidth]{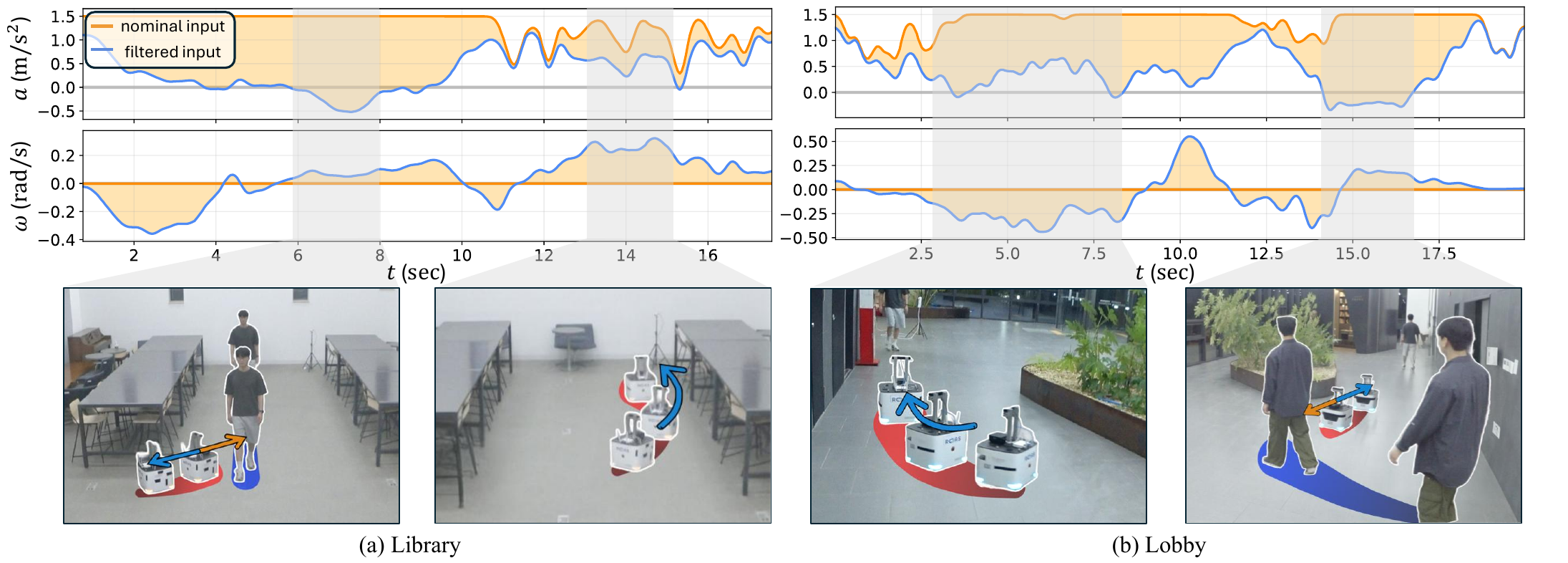 }
    \vspace{-0.2cm}

    \caption{
    \textcolor{red}{
    \textit{Safety under aggressive forward commands.}
    The nominal controller continuously commands forward motion with zero angular velocity, while the RGB-only safety filter modifies the commands in response to obstacles.
    (a) Library: the filter stops the robot to yield to a moving human (left) and steers left to avoid a static obstacle (right).
    (b) Lobby: the filter steers right to avoid the wall (left) and stops to wait for a moving human to pass (right).
    }}
    \label{fig:joystick_real_exp}
    \vspace{-0.6cm}
\end{figure*}

\subsubsection{Ablation on dataset aggregation and augmentation}

We ablate two ingredients of the distillation dataset: aggregating the rollouts of all three nominal policies rather than training a separate student per nominal, and action augmentation of the teacher labels.
We train each variant and evaluate the filters in the simulation using same conditions and trials as in Table~\ref{tab:sim_results}.
Table~\ref{tab:ablation} shows that combining both raises the mean success \textcolor{red}{and collision-free} rate over all scenes and nominal policies to 74.0\% \textcolor{red}{and 86.1\%}, close to the teacher's 76.8\% \textcolor{red}{and 91.5\%}, whereas either ingredient alone does not improve over the per-nominal, non-augmented student.
This indicates that the two are complementary.
Augmentation broadens the coverage of the nominal action space, and aggregation supplies the diverse visited states, yielding a single filter that transfers across nominal policies.

\section{Hardware Experiments}
To further validate our RGB-only safety filter in real-world dynamic scenarios, we conduct hardware experiments across the target environments and diverse nominal policies. We assess the resulting safety improvement by measuring navigation performance and analyzing failure cases, and by examining the filter's response to aggressive nominal inputs.

\subsection{Hardware Platform}
We use a differential wheeled robot for real-world experiments. The robot is equipped with a forward-facing ZED2 camera for RGB observations and an NVIDIA Jetson Orin NX for onboard inference. The safety filter runs at 9.2 ms per inference on the onboard computer.  
For nominal policies that require a relative goal position, we obtain it from wheel odometry, which is not provided to the safety filter.

\subsection{Navigation Results}
We evaluate navigation performance over 10 episodes for each combination of nominal policy, filter setting, and environment, resulting in a total of 180 episodes. Five humans who do not appear in the training data participate as dynamic obstacles. Table~\ref{tab:real_results} shows that our RGB-only safety filter improves success rates \textcolor{red}{and collision-free rates}, and increases clearance from obstacles across nominal policies, which is consistent with the simulation results. The failure case analysis reveals that the proposed filter effectively reduces collisions with both static and dynamic obstacles in most cases, while it occasionally increases timeout failures, as the filter reduces speed in densely occupied regions. Fig.~\ref{fig:thumbnail} shows a qualitative example in which the filtered RL policy detours around a dynamic obstacle, whereas the nominal policy alone collides with it. These results indicate that a safety filter distilled entirely in simulation transfers to the real world and improves the safety of RGB-only navigation policies.

\subsection{Safety Under Aggressive Nominal Inputs}
\textcolor{red}{
To examine the behavior of the safety filter under aggressive commands, we apply a persistent forward nominal input without obstacle avoidance. 
The target linear velocity is fixed at $v_{\mathrm{target}}=0.75~\mathrm{m/s}$, and the nominal acceleration is computed as $a_{\mathrm{nom}}=(v_{\mathrm{target}}-v_{\mathrm{cur}})/\Delta t$ with $\Delta t=0.1~\mathrm{s}$, while the nominal angular velocity is fixed at $\omega_{\mathrm{nom}}=0$. 
Thus, the nominal command drives the robot straight ahead regardless of surrounding obstacles.
Fig.~\ref{fig:joystick_real_exp} shows that the proposed filter overrides these commands when necessary.
In the Library, the robot slows down and waits for a crossing human, while steering away from a static obstacle instead of continuing straight. In the Lobby, the filter similarly turns away from the wall and stops to yield to a moving human. 
These results demonstrate that our filter can generate appropriate braking and steering interventions even under persistently unsafe nominal commands.
}

\section{Conclusion}

In this paper, we presented an RGB-only safety filter for visual navigation in dynamic environments.
We used a photorealistic simulator of the target environment to provide privileged state that is unavailable at real-world deployment to a teacher CBF filter, and designed the teacher to account for static obstacles and human velocity uncertainty while keeping its decisions inferable from the student's RGB observations.
Distilling this teacher into a student yields a safety filter that requires no robot localization, obstacle state estimation, or 3D scene representation at deployment.
Simulation and real-world experiments show that the distilled filter improves the success and collision-free rate of diverse nominal policies and enlarges the clearance to both static obstacles and the human. 
In future work we aim to scale training across scenes toward a single RGB-only safety filter that transfers to unseen environments and nominal policies.


\bibliographystyle{IEEEtranBST/IEEEtran}
\bibliography{reference, IEEEtranBST/IEEEabrv}

\end{document}